\documentclass[10pt,twocolumn,letterpaper]{article}

\usepackage{cvpr}              

\usepackage{graphicx}
\usepackage{amsmath}
\usepackage{amssymb}
\usepackage{booktabs}

\usepackage[pagebackref,breaklinks,colorlinks]{hyperref}

\usepackage[capitalize]{cleveref}
\crefname{section}{Sec.}{Secs.}
\Crefname{section}{Section}{Sections}
\Crefname{table}{Table}{Tables}
\crefname{table}{Tab.}{Tabs.}

\def\cvprPaperID{*****} 
\def\confName{CVPR}
\def\confYear{2022}

\begin{document}

\title{Exploring adaptation of VideoMAE for Audio-Visual Diarization \& Social @ Ego4d Looking at me Challenge}

\author{Yinan He\\
Beijing University of Posts and Telecommunications\\
{\tt\small heyinan@bupt.cn} \\
\\
Guo Chen\\
State Key Lab for Novel Software Technology, Nanjing University\\
{\tt\small chenguo1177@gmail.com} \\
}
\maketitle

\begin{abstract}
   In this report, we present the transferring pretrained video mask autoencoders(VideoMAE) to egocentric tasks for Ego4d Looking at me Challenge. VideoMAE is data-efficient pretraining model for self-supervised video pre-training and can easily transfer to downstream tasks. We show that the representation transferred from VideoMAE has good Spatio-temporal modeling and the ability to capture small actions. We only need to use egocentric data to train 10 epochs based on VideoMAE which pretrained by the ordinary videos acquired from a third person's view, and we can get better results than baseline on Ego4d Looking at me Challenge.
\end{abstract}

\section{Introduction}
\label{sec:intro}

Ego4d~\cite{ego4d} has organized and published a large-scale, widely distributed egocentric video dataset. It makes many first-person perceptual tasks possible and promotes the development of related fields.

In human communication, "look at me" is an indispensable action to interact with others. In 2022Ego4D challenge@ECCV, we were asked to identify the communicative acts towards the camera-wearer. Specifically, given a video, the face of the social partner has been localized and identified, and each visible face should be classified as to whether is looking at the camera-wearer or not.

To alleviate this challenge, we take a face sequence as our model input and get the probability of whether this face in the clip looks at the camera-wearer. We replace the LSTM model with VideoMAE~\cite{videomae} on the official baseline, and simply fine-tune the videoMAE pretrained on Something-Something-V2~\cite{sthsth} with Ego4d annotations. \textit{The code is available at \url{https://github.com/yinanhe/ego4d-look-at-me}.}


\section{Approach}
Masked Autoencoders is a new unsupervised pre-training paradigm. It is to mask the input part and achieve the pre-training result by restoring and reconstructing the mask part. MAE~\cite{mae} has been proved to be a strong competitor of pre-training methods widely used in computer vision. The pre-training model based on MAE has achieved very high fine-tuning accuracy on a wide range of vision tasks of different types and complexity. Video masked autoencoders(VideoMAE)~\cite{videomae} uses self-supervised video pre-training based on a video mask tube, and shows an efficient learning method under a small amount of data. 

Look at me aims to classify the given face in the sequence. The sequence $V$ of face $F_n$ with $T$ frames sampled in stride $s$ is denoted as $ V_{F_n} = \{{I_t^{F_n}}\ t \in [1,T] \} $, where $I$ is the face image cropped according to face region. We train the model by cropped face sequences $V$ and their annotations. We obtain feature representation by fine-tuning VideoMAE on the Ego4d dataset, to take advantage of its powerful transfer capability. Then features are input to a linear layer to predict 2 categories.

\section{Experiments}
\subsection{Experimental Configuration}
\textbf{Dataset.} 
For our experiments, we use the Ego4D~\cite{ego4d} Look-at-me benchmark. The dataset for this challenge is composed of 38242 clips. In the dataset, the data that is not looking at me is ten times more than the data that is looking at me. We sample face sequences from clips with $stride=13$ during training, and $stride=1$ during test. We extract three frames before and after the current frame as the input of the sequence model in each step. The blank frame will be used as a fill when the interval frame does not exist.

\textbf{Training details.}
It is difficult to capture the changes in the eyes because they are very small. We use VideoMAE pretrained on Something-Something-V2 as our backbone. Finetuning settings are shown in Table.~\ref{hyparam}.
\begin{table}[]
\caption{\label{hyparam} Fine-tuning settings of VideoMAE.}
\begin{tabular}{lcc}
\hline
                                                  & VideoMAE-B                    & VideoMAE-L                    \\\hline
Resolution & \multicolumn{2}{c}{224}  \\
Optimizer                                         & \multicolumn{2}{c}{AdamW~\cite{adamw}} \\
Momentum                                          & \multicolumn{2}{c}{$\beta_1, \beta_2=0.9, 0.999$}            \\
Weight decay                                      & \multicolumn{2}{c}{0.05}                                     \\
Learning rate schedule                            & \multicolumn{2}{c}{cosine~\cite{cosine}}                             \\
Start learning rate                               & \multicolumn{2}{c}{1e-6}                                     \\
End learning rate                                 & \multicolumn{2}{c}{1e-6}                                     \\
Batch size                                        & 512                          & 256                           \\
Learning rate                                     & 5e-6                         & 4e-6                          \\
Warmup epoch & \multicolumn{2}{c}{3}                                        \\
Epochs                             & \multicolumn{2}{c}{10}                                       \\
Scale                                             & \multicolumn{2}{c}{{[}0.08, 1.00{]}}                         \\
Jitter aspect ratio                               & \multicolumn{2}{c}{{[}0.75, 1.33{]}}                         \\
Color jitter                                      & \multicolumn{2}{c}{0.4}                                      \\
Rand augment~\cite{cubuk2020randaugment}                                     & \multicolumn{2}{c}{rand-m7-n4-mstd0.5-inc1}                  \\ \hline
\end{tabular}
\end{table}

\subsection{Results}

In Table.~\ref{result}, we evaluate our fine-tuning VideoMAE in the Ego4D Look at me benchmark and compare it with the challenge baseline: LSTM~\cite{LSTM}. Our results show that VideoMAE-L outperforms the baseline by $7\%$ mAP and $19\%$ accuracy.

\begin{table}[]
\centering
\caption{\label{result} Results of Experiments.}
\begin{tabular}{lll}
\hline
           & mAP  & Acc  \\ \hline
Baseline   & 0.66 & 0.74 \\
VideoMAE-B & 0.69 & 0.91 \\
VideoMAE-L & 0.73 & 0.93 \\ \hline
\end{tabular}
\end{table}

\section{Conclusion}
We implement VideoMAE as a pre-training model in a simple and efficient way for fine-tuning in egocentric video, which is used to perceive the interaction relationship of people in egocentric. Our methods have achieved competitive performance in the Look at me challenge. This method can make good use of the existing public video pre-training model, and easily transfer the representation learned in the third-person video to the egocentric perspective task.

\textbf{Limitaions.} Since we do not add more tricks to the training strategy, our method is vulnerable to the impact of data imbalance.
{\small
\bibliographystyle{ieee_fullname}
\bibliography{egbib}
}

\end{document}